\documentclass{article}
\usepackage{spconf,amsmath,graphicx}
\usepackage{amssymb,amsthm,mathrsfs,picture}

\graphicspath{{.}{figures/}}

\def\x{{x}}

\def\om{{\omega}}
\def\omu{{\omega_1}}
\def\omd{{\omega_2}}

\def\Leb{{L}}
\def\exp{{\mathbb E}}
\def\var{{\mathrm{Var}}}
\def\argmin[#1]{{\operatorname*{arg\min}_{#1}}}

\def\Real{{\mathbb R}}
\def\Complex{{\mathbb C}}
\def\Rot{{R}}

\def\Vol{{\mathbf{X}}}
\def\vol{{X}}
\def\volfun{{\mathcal{X}}}
\def\Im{{\mathbf{I}}}
\def\im{{I}}
\def\imfun{{\mathcal{I}}}
\def\Noise{{\mathbf{E}}}

\def\proj{{P}}
\def\projfun{{\mathcal{P}}}

\def\ctfop{{T}}
\def\ctfopfun{{\mathcal{T}}}
\def\discfun{{\mathcal{S}}}
\def\imop{{M}}
\def\imopfun{{\mathcal{M}}}
\def\volspace{{\mathscr{V}}}
\def\imspace{{\mathscr{I}}}
\def\Nres{{N_{\mathrm{res}}}}
\def\SNRhet{{\mathrm{SNR_{het}}}}

\title{Covariance estimation using conjugate gradient for 3D classification in Cryo-EM}
 \name{Joakim And\'{e}n$^{\star}$ \qquad Eugene Katsevich$^{\dagger}$ \qquad Amit Singer$^{\star}$\thanks{This research was partially supported by Award Number R01GM090200 from the NIGMS and Award Number LTR DTD 06-05-2012 from the Simons Foundation.}}

 \address{$^{\star}$ Program in Applied and Computational Mathematics, Princeton University, Princeton, NJ \\
     $^{\dagger}$ Department of Statistics, Stanford University, Stanford, CA}

\begin{document}
%\ninept
%
\maketitle
\begin{abstract}
Classifying structural variability in noisy projections of biological
macromolecules is a central problem in Cryo-EM. In this work, we build on a
previous method for estimating the covariance matrix of the three-dimensional
structure present in the molecules being imaged. Our proposed method allows for
incorporation of contrast transfer function and non-uniform distribution of
viewing angles, making it more suitable for real-world data.  We
evaluate its performance on a synthetic dataset and an experimental dataset
obtained by imaging a 70S ribosome complex.
\end{abstract}
\section{Introduction}
\label{sec:intro}

A variety of techniques exist to estimate the structure of biological
macromolecules: X-ray crystallography, nuclear magnetic resonance (NMR)
spectroscopy, and cryo-electron microscopy (Cryo-EM).  While X-ray
crystallography provides the best resolution, it requires
crystallization -- a challenging task for biological molecules. NMR
spectroscopy is limited in only being suitable for small molecules
($<50~\mathrm{kDa}$). Cryo-EM does not require crystallization of the molecules
and recent advances in detector technology enable structure determination at
near-atomic resolution for sizes greater than $200~\mathrm{kDa}$
\cite{Bai2014,resolution_revolution}. However, biological molecules are prone
to radiation damage, so the electron dose is limited ($5$--$20$
$\mathrm{e}^-/\mathrm{\AA^2}$), resulting in images with poor signal-to-noise
ratio (SNR). Dealing with high noise levels is therefore crucial to any Cryo-EM
analysis.

In this work, we consider single particle reconstruction (SPR) Cryo-EM. Here,
we assume that identical copies of a single molecule are rapidly frozen in a
thin layer of vitreous ice and kept frozen during imaging. Each copy then has
a random orientation and position. To reconstuct the molecular structure from
these noisy projections, we first estimate the Euler angles describing each orientation.  Classical tomographic inversion methods are then applied to obtain
a three-dimensional voxel structure, or ``volume,'' representing the molecule.
Using this volume, Euler angles are re-estimated, and the process is repeated
until convergence. This scheme, known as iterative refinement, can be
implemented using a variety of algorithms at each step
\cite{frank2006three,van2000single}.

Most such algorithms assume that all molecules imaged have the same structure.
However, this assumption is often invalid since many molecules exist in a
multitude of states. Modeling of multiple molecular states in a Cryo-EM
dataset is known as the heterogeneity problem and has attracted much attention
in recent years. 

One proposed solution is to model the distribution of images using a projected
mixture model \cite{sigworth2010chapter}. This approach has proven successful,
but is computationally very expensive and requires the number of
states to be known in advance. Other methods rely on the fact that Fourier
transforms of projections arising from the same volume coincide on a single
line, known as the common line. Images can then be clustered by measuring
correlations between lines in Fourier space \cite{hk_08,shat10}.

Often, molecular states will differ only locally, so estimation of Euler angles
can be performed by fitting a single-state model. These are sufficiently
accurate for a first reconstruction. Once images have been clustered according
to different molecular states, angles are re-estimated during iterative
refinement on each cluster. In the following, we assume that the angles are
known and focus on the task of classification.

The work in this paper draws on a concept introduced by Penczek et al., who
determine the volumes of different molecular states by estimating their
covariance \cite{penczek2011identifying}. Given a set of $C$ volumes, their
covariance matrix will be dominated by $C-1$ eigenvectors, or ``eigenvolumes.''
Each projected image is then approximated as a combination of projected
eigenvolumes. Clustering the resulting coordinates into $C$ classes now yields
a good classification of the images. To estimate the covariance matrix, the
authors propose a bootstrapping approach in which multiple subsets of the
dataset are used to reconstruct multiple volumes whose covariance is then
calculated. Unfortunately, this heuristic method offers no theoretical
guarantees.

Katsevich et al. have proposed an estimator for the volume covariance matrix
that remedies this problem \cite{gene}. This estimator has several useful
properties: it converges to the population covariance matrix as the number of
images goes to infinity, does not assume a particular distribution of molecular
states, and does not require knowing the number of classes $C$. Indeed, $C$ can be estimated from the spectrum of the covariance matrix.

Unfortunately, calculating this estimator involves the inversion of a
high-dimensional linear operator, making direct calculation intractable for
typical problems. To solve this, the authors replace the operator by a sparse,
block-diagonal approximation that can be more easily inverted. However, this is
only valid for a uniform distribution of viewing angles and does not
incorporate the contrast transfer function (CTF) of the microscope, which is
necessary for real-world data.

In this paper, we instead invert the original linear operator using the
conjugate gradient (CG) method. The operator can be decomposed as a sum of
sparse operators, and so applying it is computationally cheap. As a result,
the CG inversion has an overall computational complexity of $O(n\Nres^{7})$,
where $n$ is the number of images and $\Nres$ is the effective resolution of
the model. This approach also has the advantage of enabling a non-uniform
distribution of viewing angles and allows us to incorporate the effect of the
CTF, as we demonstrate through classification on both simulated and
experimental datasets.

\section{Cryo-EM Imaging Model}
\label{sec:imaging}

In this paper, we shall represent the molecular structure using its Coulomb
potential function in three dimensions, defined by some $\volfun \in
\Leb^1(\Real^3)$. The imaging process includes a lowpass filtering, so we cannot
expect to represent volumes accurately above a certain frequency. More
importantly, since our goal is classification, we only need enough resolution
to distinguish one molecular state from another. To reduce computational cost,
we therefore restrict $\volfun$ to some finite-dimensional subspace $\volspace$
of $\Leb^1(\Real^3)$ where the frequency content is concentrated in a ball of
radius $\Nres\pi/2$, yielding an effective resolution of $\Nres$. 

We can represent a particular viewing direction as an axis of integration and
an in-plane rotation, that is an element $\Rot$ of $\mathrm{SO}(3)$, the group
of orientation-preserving rotations in $\Real^3$.  The projection of $\volfun$
corresponding to the rotation $\Rot$ is then given by
\begin{equation}
\projfun\volfun(x,y) = \int_\Real \volfun(\Rot^Tr)\,dz,
\end{equation}
where $r = (x,y,z)^T$.

An electron microscope never captures the actual projection $\projfun \volfun$.
Instead, it registers a projection filtered by a CTF $H(\omega)$ which depends
on microscope optics and the wavelength of the electron beam used
\cite{frank2006three}.

Let us define the $D$-dimensional Fourier transform of a function $f \in
\Leb^1(\Real^D)$ (here $D$ is typically $2$ or $3$) as
\begin{equation}
\widehat{f}(\om) = \int_{\Real^D} f(\x) e^{-i\om ^T \x}\, d\x
\end{equation}
for any $\om \in \Real^D$. The Fourier transform of the CTF-filtered projection is then 
\begin{equation}
H(\om)\cdot\widehat{\projfun\volfun}(\om)~.
\end{equation}

Instead of applying the CTF to the filtered image, we can apply it to the
volume prior to projection. The Fourier slice theorem \cite{natterer} tells us
that
\begin{equation}
	\widehat{\projfun\volfun}(\omu,\omd) = \widehat{\volfun}(\Rot^T(\omu,\omd,0)^T)~.
\end{equation}
The CTF is radially symmetric, so it can be extended symmetrically to $\Real^3$. 
We thus have $H(\Rot^T\om) = H(\om)$, and so
\begin{equation}
H(\omu,\omd)\cdot\widehat{\projfun\volfun}(\omu,\omd) = 
(H \cdot \widehat{\volfun})(\Rot^T(\omu,\omd,0)^T)~.
\end{equation}
Letting $\ctfopfun$ denote spatial filtering by $H(\om)$, the CTF-filtered projection is then $\projfun\ctfopfun\volfun$.

CTFs have several zero-crossings, so reconstruction is not possible from a
dataset obtained from a single CTF. Experiments thus use a number of microscope
configurations, resulting in different CTFs covering the entire frequency
spectrum. 

Finally, the image is registered on a discrete grid of size $N$-by-$N$. As
mentioned previously, however, we only consider volumes of effective resolution
$\Nres$. Consequently, we restrict our images to those in a finite-dimensional
space $\imspace$ with frequency content centered in a ball of radius
$\Nres\pi/2$. The sampling operator mapping $\Leb^1(\Real^2)$ to $\imspace$ is
denoted $\discfun$.

Putting everything together, the image $\imfun$ obtained from $\volfun$
through convolution with $\ctfopfun$, projection by $\projfun$ and sampling by $\discfun$ is given by
\begin{equation}
\imfun = \discfun\projfun\ctfopfun \volfun = \imopfun \volfun~,
\end{equation} 
where we have introduced the imaging operator $\imopfun =
\discfun\projfun\ctfopfun$.

Since both $\volspace$ and $\imspace$ are of finite dimension, we can
represent them using finite bases. Let $\dim \volspace = p$ and $\dim
\imspace = q$. We can then represent $\volfun$ and $\imfun$ as vectors $\vol$
and $\im$ in $\Real^p$ and $\Real^q$, respectively. Similarly, $\projfun$ and $\ctfopfun$ have matrix representations $\proj$ and $\ctfop$, respectively, obtained by least-squares approximation. Taken together, we have the imaging matrix $\imop$. Note that $\discfun$ is no longer present since $\proj$ and $\ctfop$ already project onto a finite-dimensional space.

\section{Volume Covariance}
\label{sec:covariance}

\subsection{Covariance estimator}
\label{sec:stat}

To model the variability of volumes in the dataset, let $\Vol_s$ for $s =
1,\ldots,n$ be a collection of independent and identically distributed discrete
random variables in $\Real^p$, each taking the value $\vol_c$ with probability
$p_c$ for $c = 1,\ldots,C$. These random variables have mean $\mu_0 =
\exp[\Vol_s]$ and covariance matrix
\begin{equation}
\Sigma_0 = \var[\Vol_s] = \exp[(\Vol_s-\exp[\Vol_s]) (\Vol_s-\exp[\Vol_s])^H]~,
\end{equation}
where $u^H$ is the conjugate transpose of the vector $u$. Since $\Vol_s$ is a
discrete random variable with $C$ states, $\Sigma$ has rank $C-1$.

To estimate $\mu_0$ and $\Sigma_0$, we consider the statistics of the projected
images. Specifically, we define the random variables
\begin{equation}
\Im_s = \imop_s \Vol_s+\Noise_s~,
\end{equation}
where $\Noise_s$ are independent and identically distributed zero-mean random noise vectors, independent of
 $\imop_s$ and $\Vol_s$, with $\var[\Noise_s] = \sigma^2 I_q$. The expected value of $\Im_s$ is
\begin{equation}
\label{eq:exp-Im}
\exp[\Im_s] = \imop_s \mu_0~,
\end{equation}
while its covariance is given by
\begin{equation}
\label{eq:var-Im}
\var[\Im_s] = \imop_s \Sigma_0 \imop_s^H+\sigma^2 I_q~,
\end{equation}
where $\imop_s^H$ is the conjugate transpose of the imaging operator $\imop_s$ and
$I_q$ is the $q$-by-$q$ identity matrix.

Let us consider the realizations $\im_s$ of $\Im_s$ for $s = 1,\ldots,n$. Following \eqref{eq:exp-Im} and \eqref{eq:var-Im}, we define the
following estimators for $\mu_0$ and $\Sigma_0$:
\begin{equation}
	\label{eq:mu-est}
\mu_n = \argmin[\mu] \frac{1}{n} \sum_{s=1}^n \|I_s - \imop_s \mu\|^2~,
\end{equation}
\begin{align}
	\label{eq:sigma-est}
\Sigma_n = \argmin[\Sigma] \frac{1}{n} \sum_{s=1}^n \|&(I_s-\imop_s\mu_n)(I_s-\imop_s\mu_n)^H \\ 
\nonumber
& - (\imop_s\Sigma \imop_s^H+\sigma^2 I_q)\|_F^2~,
\end{align}
where $\|\cdot\|_F$ is the Frobenius matrix norm.

Differentiating and setting to zero in \eqref{eq:mu-est}, we get
\begin{equation}
\label{eq:mu-lhs}
A_n \mu_n = b_n~,
\end{equation}
where $A_n$ and $b_n$ are given by
\begin{equation}
A_n = \frac{1}{n}\sum_{s=1}^n \imop_s^H \imop_s,\qquad b_n = \frac{1}{n} \sum_{s=1}^n \imop_s^H I_s
\end{equation}

Applying the same process to \eqref{eq:sigma-est}, we obtain
\begin{equation}
\label{eq:sigma-lhs}
L_n(\Sigma_n) = B_n,
\end{equation}
where $L_n: \Complex^{p\times p} \rightarrow \Complex^{p\times p}$ is 
the linear operator defined by
\begin{equation}
L_n(\Sigma) = \frac{1}{n}\sum_{s=1}^n \imop_s^H \imop_s \Sigma \imop_s^H \imop_s
\end{equation}
and
\begin{align}
B_n = \frac{1}{n} \sum_{s=1}^n &\imop_s^H(I_s-\imop_s\mu_n)(I_s-\imop_s\mu_n)^H\imop_s \\
\nonumber
& - \sigma^2 \frac{1}{n} \sum_{s=1}^n \imop_s^H \imop_s~.
\end{align}

Calculating $\mu_n$ and $\Sigma_n$ thus amounts solving \eqref{eq:mu-lhs} and
\eqref{eq:sigma-lhs}. Since $\imspace$ contains images of effective resolution
$\Nres$, $q = O(\Nres^2)$. Likewise, $p = O(\Nres^3)$.  The matrix $A_n$ is
$p$-by-$p$, can therefore be na\"ively inverted with a complexity $O(\Nres^9)$.
However, if we were to compute the matrix representation of $L_n$, this would
be a $p^2$-by-$p^2$ matrix, and its inversion would take $O(\Nres^{18})$. So
while we may be able to calculate $A_n^{-1}$, inverting $L_n$ poses a much
greater challenge.

\subsection{Inversion of $L_n$}
\label{sec:inversion}

Since direct inversion of the matrix of $L_n$ is not an option, we turn to
other methods of solving \eqref{eq:sigma-lhs}. If $L_n$ can be calculated fast,
the conjugate gradient method provides an viable approach for estimating
$\Sigma_n$.

By choosing appropriate spaces of $\volspace$ and $\imspace$ and equipping these with
well-behaved bases, $\proj_s$ can be expressed as a block-diagonal matrix
consisting of $O(\Nres)$ blocks of size $O(\Nres)$-by-$O(\Nres^2)$. The
application of the CTF, $\ctfop$, is also represented by a block-diagonal 
matrix with diagonal blocks in this basis. Details on the construction
of these bases can be found in Katsevich et al. \cite{gene}. All matrix
multiplications are therefore done in blocks, reducing computational
complexity. 

The CTF matrices $\ctfop_s$ in $L_n$ result in certain frequencies being
amplified and others attenuated. Because the noise in our images $I_s$ is
white, the stability of the inversion $\om$ thus depends on $|H_s(\om)|$. We
would therefore like to only invert $L_n$ when $|H_s(\om)|$ is large on
average, corresponding to the dominant eigenvalues of $L_n$. Since the
conjugate gradient method first constructs the larger eigenvalues of the
inverse before moving on to smaller eigenvalues, the desired result is obtained
by limiting the number of iterations, yielding an implicit regularization
\cite{hanke1995conjugate}. The overall complexity is therefore that of applying
$L_n$, which can be shown to be $O(n\Nres^7)$.

\subsection{Classification}
\label{sec:cluster}

As mentioned previously, $\Sigma_0$ has
$C-1$ non-zero eigenvalues and the eigenvectors, together with $\mu_0$, define an
affine space containing all the volumes. Due to
noise, this is not the case for $\Sigma_n$, although it converges to $\Sigma_0$ as $n$
increases. In numerical experiments, we find that for large $n$, $\Sigma_n$ will
contain $C-1$ dominant eigenvalues and the associated eigenvectors
approximate the eigenvectors of $\Sigma_0$.

Assembling the dominant $C-1$ eigenvectors into a matrix $U_n$,
we can associate with each image $\im_s$ a coordinate vector $\alpha_s$ such that $\|(\mu_n+U_n\alpha)-\im_s\|^2$ is minimized.
If $\im_s$ is a projection of the volume $\vol_c$, $\mu_n+U_n\alpha_s$
should be close to $\vol_c$. As a result, the $\alpha_s$ cluster 
according to molecular state.

This lets us classify the images according to their molecular structure.
Applying a clustering algorithm to the $\alpha_s$ vectors, the images generated
by the a given volume will be found in the same cluster. We use a Gaussian mixture 
model (GMM) trained using the expectation-maximization (EM) algorithm 
\cite{dempster}. Once images are associated with a particular molecular structure, standard tomographic
inversion techniques can be applied to recover that structure.

\begin{figure}[t]
	\setlength{\unitlength}{1in}
	\begin{picture}(3.5,0.9)
	\put(-0.2,-0.1){\includegraphics[width=1.1\columnwidth]{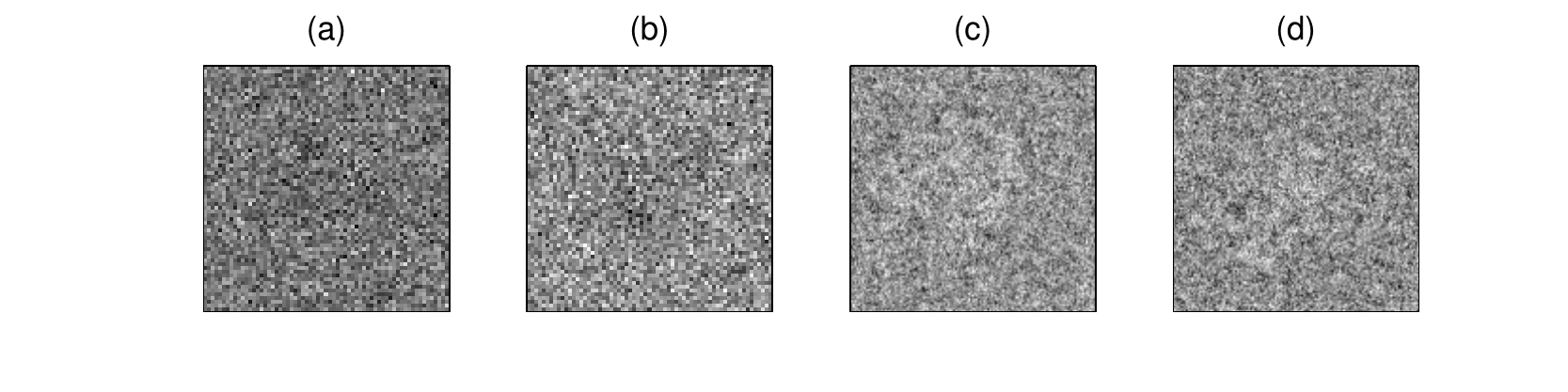}}
	\end{picture}
	\caption{\label{fig:samples} Sample projection images from the synthetic dataset (a,b) and experimental images of the 70S ribosome (c,d).}
\end{figure}

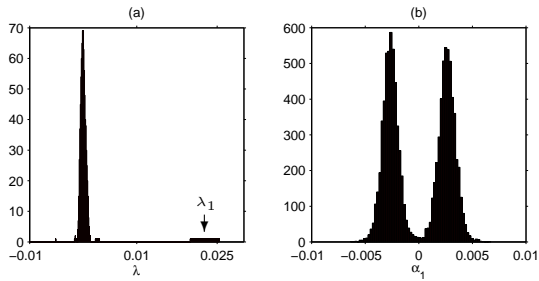
\begin{figure}[t]
	\setlength{\unitlength}{0.9in}
	\begin{picture}(3.5,1.55)
	\put(0,-0.2){\includegraphics[width=0.99\columnwidth]{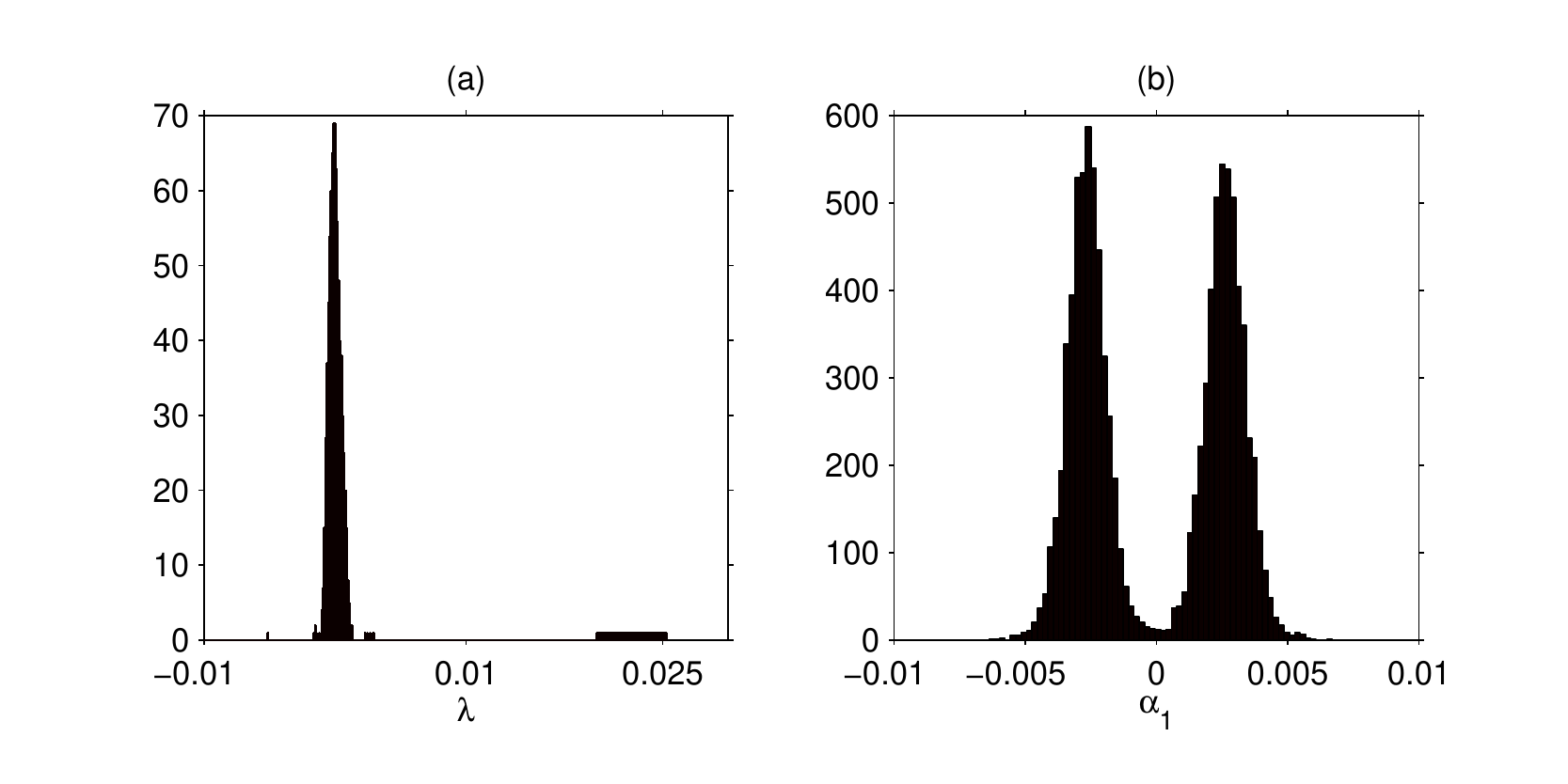}}
	\put(1.45,0.35){\tiny $\lambda_1$}
	\put(1.50,0.30){\vector(0,-1){0.1}}
	\end{picture}
	\caption{\label{fig:synth} (a) The eigenvalue histogram of $\Sigma_n$ 
	obtained from synthesized data. 
	(b) The histogram of the coordinate $\alpha_1$.}
\end{figure}

\section{Numerical Experiments}
\label{sec:numeric}

\subsection{Synthetic data}
\label{sec:synthetic}

\begin{figure}[t]
	\setlength{\unitlength}{0.9in}
	\begin{picture}(3.5,3.2)
	\put(0,-0.4){\includegraphics[width=0.99\columnwidth]{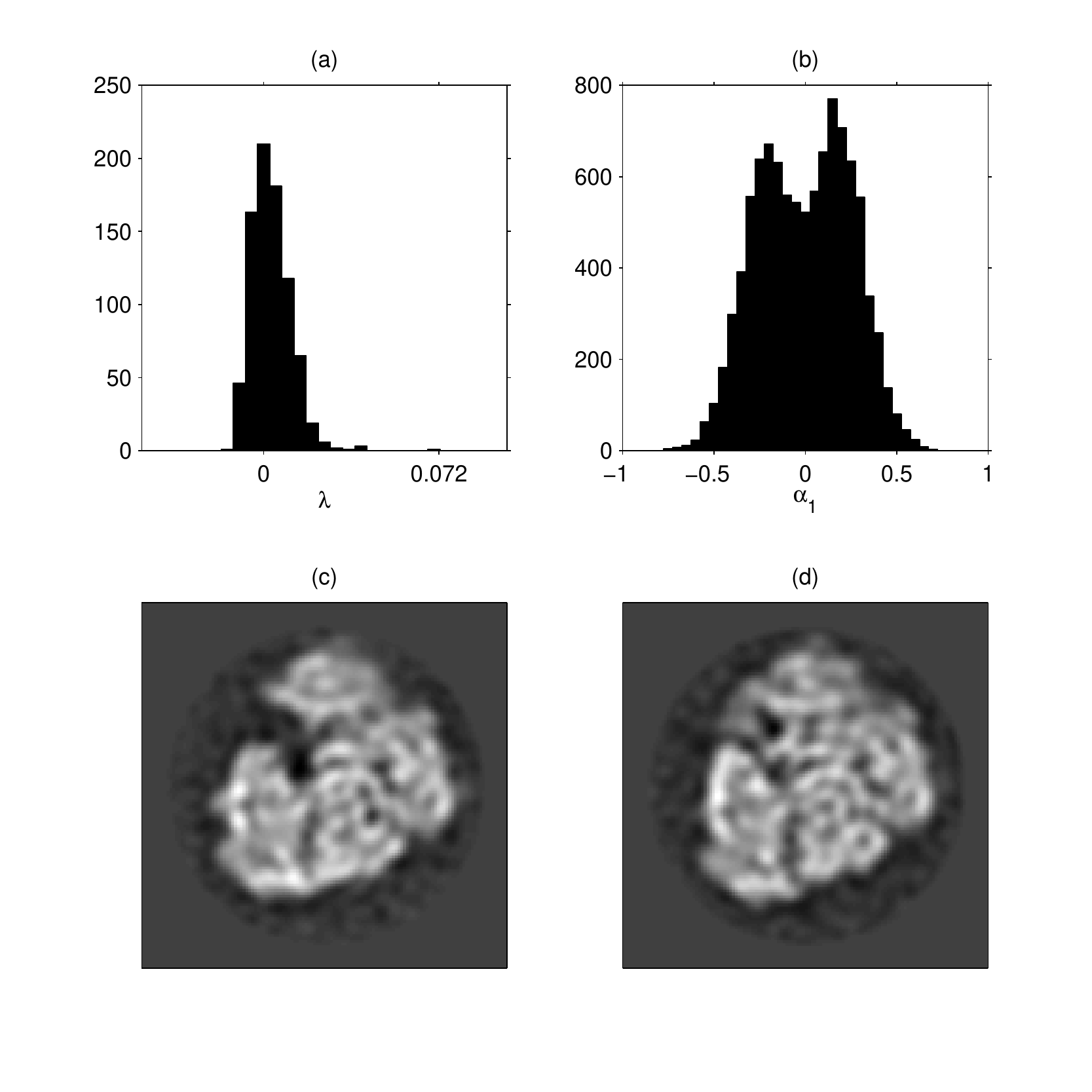}}
	\put(1.43,2){\tiny $\lambda_1$}
	\put(1.48,1.95){\vector(0,-1){0.1}}
	\end{picture}
	\caption{\label{fig:rib70s} (a) The eigenvalue histogram for $\Sigma_n$ 
	obtained from experimental images of the 70S ribosome complex. 
	(b) The histogram of the coordinate $\alpha_1$.
	(c,d) Cross-sections of estimated volumes.}
\end{figure}
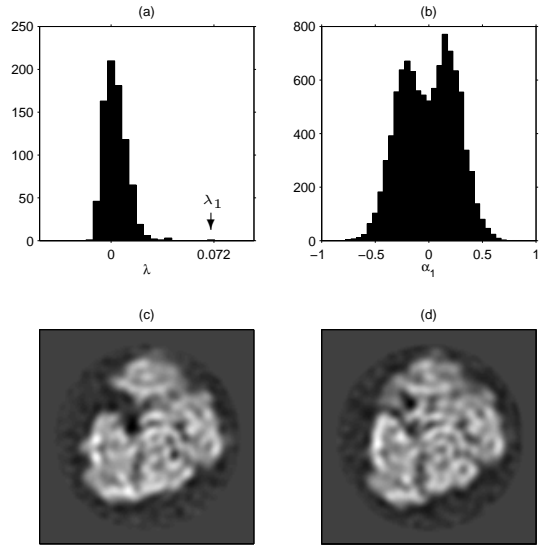

To evaluate the above method, we apply it to a synthetic dataset consisting of
$n = 10000$ images generated from $C = 2$ volumes, projected along random
viewing directions, and filtered by one of seven CTFs. Each image is sampled on
a $65$-by-$65$ grid and a Gaussian noise of variance $\sigma^2$ is added.  In
this section, we have a heterogeneous SNR $\SNRhet = 0.005$ (for a discussion
of $\SNRhet$, see \cite{gene}). Sample projections are shown in Figure
\ref{fig:samples}(a,b). 

The algorithm is run with an effective resolution of $\Nres = 17$ for $10$
iterations. The total running time is $50~\mathrm{min}$ on a $3~\mathrm{GHz}$
quad-core CPU with $4~\mathrm{GB}$ of memory. The eigenvalues of $\Sigma_n$ are
shown in Figure \ref{fig:synth}(a).  One eigenvalue is separated from the rest
by a spectral gap of $8.4$, representing the heterogeneity in the dataset.
Calculating the coordinates $\alpha_s$ of the images $\im_s$, we obtain two
well-separated distributions. Clustering the coordinates recovers the
original classes with $99.8\%$ accuracy.

\subsection{Ribosome 70S complex}
\label{sec:ribosome}

We also apply the method an experimental dataset consisting of $n =
10000$ images sampled on a $130$-by-$130$ grid. Sample images are shown in Figure \ref{fig:samples}(c,d). The projections correspond to
different molecular states of a 70S ribosomal complex from E. Coli generously
provided by J. Frank's group at Columbia University \cite{frank70s_10k}.  To
estimate orientations, the RELION software was run with one class
\cite{scheres2012relion}.

Running the algorithm with $\Nres = 17$ for $10$ iterations, we obtain
$\Sigma_n$ with the spectrum shown in Figure \ref{fig:rib70s}(a). The total
running time was $46~\mathrm{min}$.  Here, one dominant eigenvalue at
$\lambda_1 = 0.072$ is well-separated from the bulk of the spectrum by a gap of
$1.75$. We therefore conclude that the dataset contains two classes.
Calculating the coordinate $\alpha_1$, we obtain the bimodal distribution in
\ref{fig:rib70s}(b). We then cluster $\alpha_1$ and send each class to RELION
for reconstruction, which yields the cross-sections shown in Figure
\ref{fig:rib70s}(c) and (d). The volumes are differentiated in the central
cavity and in the rotation of the upper part.  Compared to the labeling
provided with the dataset, our clustering achieves an accuracy of $87\%$.

\section{Conclusion}
\label{sec:conclusion}

Replacing the sparse approximation of Katsevich et al. with an iterative
approach, we obtain a more flexible method for covariance estimation
capable of tackling experimental datasets. Compared to other
algorithms, it is less computationally intensive and provides an estimate of the number of classes, simplifying the classification of molecular structure.

\bibliographystyle{IEEEbib}
\bibliography{refs}

\end{document}